\RequirePackage{fix-cm}
\documentclass{svjour3}                     
\smartqed  
\usepackage{mathptmx}      
\journalname{Machine Learning Journal}
\usepackage{algorithm}
\usepackage{algorithmic}
\usepackage{url}
\usepackage{amsmath,amssymb}
\usepackage{graphicx}
\usepackage{subfig}
\usepackage{multirow}
\usepackage{color} 
\usepackage{cite} 
\usepackage{hyperref}
\usepackage{comment}
\usepackage{gensymb}

\DeclareMathOperator*{\argmin}{arg\,min}

\begin{document}

\title{\textbf{Coupling Matrix Manifolds Assisted Optimization for Optimal Transport Problems}}

\author{Dai Shi \and Junbin Gao \and Xia Hong \and S.T.~Boris~Choy \and Zhiyong Wang}

\institute{Dai Shi \at
              Discipline of Business Analytics, The University of Sydney Business School, The University of Sydney, NSW 2006, Australia. \\
              \email{dai.shi@sydney.edu.au}
              \and
            Junbin Gao \at
              Discipline of Business Analytics, The University of Sydney Business School, The University of Sydney, NSW 2006, Australia. \\
              \email{junbin.gao@sydney.edu.au}
              \and  
             Xia Hong \at
             Department of Computer Science, University of Reading,  Reading,  RG6 6AY,UK. \\
             \email{x.hong@reading.ac.uk}
             \and
            S. T. Boris Choy \at   Discipline of Business Analytics, The University of Sydney Business School, The University of Sydney, NSW 2006, Australia. \\
              \email{boris.choy@sydney.edu.au}  
            \and
        Zhiyong Wang \at
        School of Computer Science,  The University of Sydney, NSW 2006, Australia. \\
        \email{zhiyong.wang@sydney.edu.au}
}

\date{Received: date / Accepted: date}

\maketitle
\begin{abstract} 
Optimal transport (OT) is a powerful tool for measuring the distance between two probability distributions.  In this paper, we develop a new manifold named the coupling matrix manifold (CMM), where each point on CMM can be regarded as the transportation plan of the OT problem. We firstly explore the Riemannian geometry of CMM with the metric expressed by the Fisher information.  These geometrical features of CMM have paved the way for developing numerical Riemannian optimization algorithms such as Riemannian gradient descent and Riemannian trust region algorithms, forming a uniform optimization method for all types of OT problems. The proposed method is then applied to solve several OT problems studied by previous literature. The results of the numerical experiments illustrate that the optimization algorithms that are based on the method proposed in this paper are comparable to the classic ones, for example the Sinkhorn algorithm, while outperforming other state-of-the-art algorithms without considering the geometry information, especially in the case of non-entropy regularized optimal transport.

\end{abstract}

\section{Introduction}

An Optimal Transport (OT) problem can be briefly described as to find out the optimized transport plan (defined as transportation polytope) between two or more sets of subjects with certain constraints \cite{PeyreCuturi2019}. It was firstly formalized by French mathematician Gaspard Monge in 1781\cite{monge1781memoire}, and was generalized by Kantorovich who provided a solution of Monge's problem in 1942 \cite{kantorovich1942translocation} and established its importance to logistics and economics. 

As the solution of the OT problem provides the optimized transportation plan between probability distributions, and the advance in computer science allows us to perform a large amount of computation in a high dimensional space, the optimized distance, known as the Wasserstein distance \cite{PanaretosZemel2019},  Monge-Kantorovich distance \cite{Brezis2018} and Earth Mover's distance \cite{RubnerTomasiGuibas2000}, has been treated as a target being analyzed in various aspects such as image processing \cite{rabin2015convex,FerradansPapadakisPeyreAujol2014}, pattern analysis \cite{ZhaoZhou2018,Cuturi2013,DBLP:journals/corr/MillerL16} and domain adaption \cite{courty2016optimal,MamanYairEytanTalmon2019,DBLP:journals/corr/abs-1906-00616}. 

The OT-based method for comparing two probability densities and generative models are vital in machine learning research where data are often presented in the form of point clouds, histograms, bags-of-features, or more generally, even manifold-valued data set. In recent years, there has been an increase in the applications of the OT-based methods in machine learning. The authors of  \cite{BousquetGellyTolstikhinSimon-GabrielSchoelkopf2017} approached OT-based generative modeling, triggering fruitful research under the variational Bayesian concepts, such as Wassertein GAN \cite{ArjovskyChintalaBottou2017,GulrajaniAhmedArjovskyDumoulinCourville2017}, Wasserstein Auto-encoders \cite{TolstikhinBousquetGellySchoelkopf2018,ZhangGaoJiaoLiuWangYang2019}, and Wasserstein variational inference \cite{AmbrogioniGueclueGuecluetuerkHinneMarisGerven2018} and their computationally efficient sliced version \cite{KolouriPopeMartinRohde2019}. Another reason that OT gains its popularity is convexity. As the classic Kantorovich OT problem is a constrained linear programming problem or a convex minimization problem where the minimal value of the transport cost objective function is usually defined as the divergence/distance between two distributions of loads \cite{PeyreCuturi2019}, or the cost associated with the transportation between the source subjects and targets. 
Therefore, the convex optimization plays an essential role in finding the solutions of OT. The computation of the OT distance can be approached in principle by interior-point methods, and one of the best is from \cite{LeeSidford2014}. 

Although the methods for finding the solutions of OT have been widely investigated in the literature, one of the major problems is that these algorithms are excessively slow in handling large scale OT problems. Another issue with the classic Kantorovich OT formulation is that its solution plan merely relies on a few routes as a result of the sparsity of optimal couplings, and therefore fails to reflect the practical traffic conditions. 
These issues limit the wider applicability of OT-based distances for large-scale data within the field of machine learning until a regularized transportation plan was introduced by Cuturi \cite{Cuturi2013} in 2013. By applying this new method (regularized OT), we are not only able to reduce the sparsity in the transportation plan, but also speed up the Sinkhorn algorithm with a linear convergence\cite{Knight2008}.

By offering a unique solution, better computational stability compared with the previous algorithms and being underpinned by the Sinkhorn algorithm, the entropy regularization method has successfully delivered OT approaches into modern machine learning aspects\cite{Villani2009}, such as unsupervised learning using Restricted Boltzmann Machines \cite{MontavonMuellerCuturi2016},  Wasserstein loss function \cite{FrognerZhangMobahiAraya-PoloPoggio2015}, computer graphics \cite{SolomonGoesPeyreCuturiButscherNguyenDuGuibas2015} and discriminant analysis \cite{FlamaryCuturiCourtyRakotomamonjy2018}. Other algorithms that aim for high calculation speed in the area of big data have also been explored, such as the stochastic gradient-based algorithms \cite{GenevayCuturiPeyreBach2016} and fast methods to compute Wasserstein barycenters \cite{CuturiDoucet2014}. Altschuler \emph{et al.} \cite{AltschulerWeedRigollet2017} proposed the Greenkhorn algorithm, a greedy variant of the Sinkhorn algorithm that updates the rows and columns which violate most of the constraints.

In order to meet the requirements of various practical situations, many works have been done to define suitable regularizations. For newly introduced regularizations, Dessein \emph{et al.} \cite{DesseinPapadakisRouas2018} extended the regularization in terms of convex functions. To apply OT to power functions, the Tsallis Regularized Optimal Transport (trot) distance problem was introduced in \cite{su2017order}. Furthermore, in order to involve OT into series data, the order-preserving Wassertein distance with its regularizor was developed in\cite{courty2016optimal}. In addition, to maintain the locality in OT-assisted domain adaption, the Laplacian regularization was also proposed in \cite{courty2016optimal}. While entropy-based regularizations have achieved great success in terms of calculation efficiency, those problems without such regularization are still challenging. For example, to solve a Laplacian regularized OT problem, Courty \emph{et al.} proposed a generalized conditional gradient algorithm, which is a variant of the classic conditional gradient algorithm \cite{Bertsekas1999}. In this paper, we shall compare the experimental results of several entropy and non-entropy regularized OT problems based on previous studies and the new manifold optimization algorithm proposed in Section \ref{Sec:4}. 

Non-entropy regularized OT problems arise the question about the development of a uniform and generalized method that is capable of efficiently and accurately calculating all sort of regularized OT problems. To answer this question, we first consider that all OT problems are constrained optimization problems on the transport plane space, namely the set of polytope\cite{PeyreCuturi2019}. Such constrained problems can be regarded as the unconstrained problem on a specific manifold with certain constraints. The well-defined Riemannian optimization can provide better performance than the original constrained problem with the advantage of treating lower dimensional manifold as a new search space. Consequentially, those fundamental numerical iterative algorithms, such as the Riemannian gradient descent (RGD) and Riemannian trust region (RTR), can naturally  solve the OT problems, achieving convergence under mild conditions. 

The main purpose of this paper are to propose a manifold based framework for optimizing the transportation polytope for which the related Riemannian geometry will be explored. The ``Coupling Matrix Manifold'' provides an innovative method for solving OT problems under the framework of manifold  optimization. The research on the coupling matrix manifold has rooted in our earlier paper \cite{SunGaoHongMishraYin2016} in which the so-called multinomial manifold was explored in the context of tensor clustering. The optimization on multinomial manifolds has successfully been applied to several density learning tasks  \cite{HongGao2015,HongGaoChenZia2015,HongGao2018}. More recently, Douik and Hassibi \cite{DouikHassibi2018} explored the manifold geometrical structure and the related convex optimization algorithms on three types of manifolds constructed by three types of matrices, namely the doubly stochastic matrices, symmetric stochastic matrices and positive stochastic matrices. The CMM introduced in this paper can be regarded as the generalization of their doubly positive stochastic manifolds. According to the mathematical and experimental results, the CMM framework unifies all types of OT solutions, providing closed form solutions compared with previous literature with higher efficiency, thus opening the door of solving OT problems under the manifold optimization framework.

In summary, the main contribution of this paper are three fold.
\begin{enumerate}
    \item We define the Coupling Matrix Manifold. We explore all the geometry properties of this manifold, including its tangent space, the projection mapping onto the tangent space, a numerically efficient retraction mapping and the calculation of Riemann gradient and Riemann Hessian on the manifold.
    \item Following the framework of optimization on manifolds, we formulate the Riemann optimization algorithm on the Coupling Matrix Manifold, so that most OT related optimization problems can be solved in a consistent way.
    \item We compare the newly presented algorithm with the existing algorithms in literature for several state-of-the-art OT models.
\end{enumerate}

The remainder of the paper is organized as follows. Section \ref{Sec:2} introduces CMM and its Riemannian geometry,including the tangent space, Riemannian gradient, Riemannian Hessian, and Retraction operator, all the ingredients for the Riemannian optimization algorithms.
In Section \ref{Sec:3}, we review several OT problems with different regularizations from other studies. These regularization problems will be then converted into the optimization problem on CMM so that the Riemannian version of optimization algorithms (RGD and RTR) can be applied. In Section \ref{Sec:4}, we will conduct several numerical experiments to demonstrate the performance of the new Riemannian algorithms and compare the results with classic algorithms (i.e. Sinkhorn algorithm). Finally Section \ref{Sec:5} concludes the paper with several recommendations for future research and applications.  

\section{Coupling Matrix Manifolds--CMM} \label{Sec:2}
In this section, we introduce the CMM and Riemannian geometry of this manifold in order to solve any generic OT problems \cite{PeyreCuturi2019} under the framework of CMM optimization \cite{AbsilMahonySepulchre2008}.

Throughout this paper, we use a bold lower case letter for a vector $\mathbf x\in \mathbb{R}^d$, a bold upper case letter for a matrix $\mathbf X\in\mathbb{R}^{n\times m}$, and a calligraphy letter for a manifold $\mathcal{M}$. The embedded matrix manifold $\mathcal{M}$ is a  smooth subset of vector space $\mathcal{E}$ embedded in the matrix space $\mathbb{R}^{n\times m}$. For any $\mathbf X\in\mathcal{M}$, $T_{\mathbf X}\mathcal{M}$ is the tangent space of the manifold $\mathcal{M}$ at $\mathbf X$ \cite{AbsilMahonySepulchre2008}. $\mathbf{0}_d$ and $\mathbf{1}_d \in\mathbb{R}^d$ are the $d$-dimensional vectors of zeros and ones, respectively, and $\mathbb{R}^{n\times m}_+$ is the set of all $n\times m$ matrices with real and positive elements.

\subsection{The Definition of a Manifold}
 
\begin{definition}
Two vectors $\mathbf p\in\mathbb{R}^n_+$ and $\mathbf q\in\mathbb{R}^m_+$ are coupled if $\mathbf p^T\mathbf{1}_n = \mathbf q^T\mathbf{1}_m$.  A matrix $\mathbf X\in\mathbb{R}^{n\times m}_+$ is called a coupling matrix for the coupled vectors $\mathbf p$ and $\mathbf q$ if $\mathbf X \mathbf{1}_m = \mathbf p$ and $\mathbf X^T\mathbf{1}_n = \mathbf q$. The set of all the coupling matrices for the given coupled $\mathbf p$ and $\mathbf q$ is denoted by
\begin{align}
\mathbb{C}^m_n(\mathbf p, \mathbf q) = \{\mathbf X\in\mathbb{R}^{n\times m}_+: \mathbf X \mathbf{1}_m = \mathbf p \text{ and } \mathbf X^T\mathbf{1}_n = \mathbf q\}. \label{Eq1}
\end{align}
\end{definition}

\begin{remark} The coupling condition 
\begin{equation}
 \mathbf p^T\mathbf{1}_n = \mathbf q^T\mathbf{1}_m
\end{equation}
is vital in this paper as this condition ensures a non-empty transportation polytope so that the manifold optimization process can be naturally employed. This condition is checked in Lemma 2.2 of \cite{de2014combinatorics}, and the proof of this lemma is based on the north-west corner rule algorithm described in \cite{queyranne2009multi}.
\end{remark}

\begin{remark}
The defined space $\mathbb{C}^m_n(\mathbf p, \mathbf q)$ is a subset of the classic transport plan space (or polytope) 
$$
\mathbb{P}^m_n(\mathbf p, \mathbf q) = \{\mathbf X\in\mathbb{R}^{n\times m}: \mathbf X \mathbf{1}_m = \mathbf p \text{ and } \mathbf X^T\mathbf{1}_n = \mathbf q\},
$$
where each entry of a plan $\mathbf X$ is nonnegative. In practice, this constraint on  $\mathbb{C}^m_n(\mathbf p, \mathbf q)$ does prevent the solution plan from being sparsity.  
\end{remark}

\begin{proposition}The subset $\mathbb{C}^m_n(\mathbf p, \mathbf q)$ forms a smooth manifold of dimension $(n-1)(m-1)$ in its embedding space $\mathbb{R}^{n\times m}_+$, named as the Coupling Matrix Manifold.
\end{proposition}
\begin{proof}Define a mapping $F: \mathbb{R}^{n\times m}_+ \rightarrow \mathbb{R}^{n+m}$ by
\[
F(\mathbf X) = \begin{bmatrix} \mathbf X\mathbf 1_m -\mathbf p \\ \mathbf X^T\mathbf 1_n - \mathbf q\end{bmatrix}.
\]
Hence
\[
\mathbb{C}^m_n(\mathbf p, \mathbf q) = F^{-1}(\mathbf 0_{n+m}).
\]
Clearly $DF(\mathbf X)$ is a linear mapping from $\mathbb{R}^{n\times m}_+$ to $\mathbb{R}^{n+m}$ with
\[
DF(\mathbf X)[\Delta\mathbf X] = \begin{bmatrix} \Delta\mathbf X\mathbf 1_m \\ \Delta\mathbf X^T\mathbf{1}_n\end{bmatrix}.
\]
Hence the null space of $DF(\mathbf X)$ is 
\[
\mathbf K = \{\Delta\mathbf X:  \Delta\mathbf X\mathbf{1}_m = \mathbf 0_n, \Delta\mathbf X^T\mathbf 1_n = \mathbf 0_m\}.
\]
As there are only $n+m-1$ linearly independent constraints among $\Delta\mathbf X\mathbf 1_m = \mathbf 0_n,$ and $\Delta\mathbf X^T\mathbf 1_n = \mathbf 0_m$, the rank of the null space is $nm - n - m +1 = (n-1)(m-1)$. Hence the dimension of the range will be $n+m-1$. According to the sub-immersion theorem (Proposition 3.3.4 in \cite{AbsilMahonySepulchre2008}),  
the dimension of the manifold $\mathbb{C}^m_n(\mathbf p, \mathbf q)$ is $(n-1)(m-1)$.  

This completes the proof.
\end{proof}

\noindent Several special cases of the coupling matrix manifolds that have been explored recently are as follows:

\begin{remark}
When both $\mathbf p$ and $\mathbf q$ are discrete distributions, i.e., $\mathbf p^T\mathbf{1}_n = \mathbf q^T\mathbf{1}_m = 1$ which are naturally coupled. In this case, we call $\mathbb{C}^m_n(\mathbf p, \mathbf q)$ the double probabilistic manifold, denoted by
\begin{align*}
\mathbb{P}^m_n(\mathbf p, \mathbf q) =\{\mathbf X\in\mathbb{R}^{n\times m}_+:& \mathbf X \mathbf{1}_m = \mathbf p, \mathbf X^T\mathbf{1}_n = \mathbf q,\\ &\text{ and }\mathbf p^T\mathbf{1}_n = \mathbf q^T\mathbf{1}_m = 1\}.
\end{align*}
\end{remark}

\begin{remark}The doubly stochastic multinomial manifold \cite{DouikHassibi2018}: This manifold is the special case of $\mathbb{C}^m_n(\mathbf p, \mathbf q)$ with $n=m$ and $\mathbf p=\mathbf q = \mathbf{1}_n$, e.g. 
\[
\mathbb{D}\mathbb{P}_n =\{\mathbf X\in\mathbb{R}^{n\times n}_+: \mathbf X \mathbf{1}_n = \mathbf{1}_n, \mathbf X^T\mathbf{1}_n = \mathbf{1}_n\}.
\]
\end{remark}
$\mathbb{D}\mathbb{P}_n$ can be regarded as the two-dimensional extension of the multinomial manifold introduced in \cite{SunGaoHongMishraYin2016}, defined as
\[
\mathbb{P}^m_n = \{\mathbf X\in\mathbb{R}^{n\times m}_+: \mathbf X \mathbf{1}_m = \mathbf{1}_n\}.
\]

\subsection{The Tangent Space and Its Metric}
From now on, we only consider the coupling matrix manifold $\mathbb{C}^m_n(\mathbf p, \mathbf q)$ where $\mathbf p$ and $\mathbf q$ are a pair of coupled vectors. For any coupling matrix $\mathbf X\in \mathbb{C}^m_n(\mathbf p, \mathbf q)$, the tangent space $T_{\mathbf X}\mathbb{C}^m_n(\mathbf p, \mathbf q)$ is given by the following proposition. 
\begin{proposition}\label{Prop1}The tangent space $T_{\mathbf X}\mathbb{C}^m_n(\mathbf p, \mathbf q)$ can be calculated as
\begin{align}
T_{\mathbf X}\mathbb{C}^m_n(\mathbf p, \mathbf q)=\{\mathbf Y\in\mathbb{R}^{n\times m}:  \mathbf Y\mathbf{1}_m = \mathbf{0}_n,\; \mathbf Y^T\mathbf{1}_n = \mathbf{0}_m\}
\end{align}
and its dimension is $(n-1)(m-1)$.
\end{proposition}
\begin{proof}
It is easy to prove Proposition \ref{Prop1} by differentiating the constraint conditions. We omit this. 

Also it is clear that $\mathbf Y\mathbf{1}_m = \mathbf{0}_n$ and $\mathbf Y^T\mathbf{1}_n = \mathbf{0}_m$ consist of $m+n$ equations where only $m+n-1$ conditions are in general independent because $\sum_{ij}Y_{ij} = \mathbf 1^T_n \mathbf Y \mathbf 1_m = 0$. Hence the dimension of the tangent space is $nm-n-m+1=(n-1)(m-1)$. The proof is completed.
\end{proof}

Following \cite{SunGaoHongMishraYin2016,DouikHassibi2018}, we still use the Fisher information as the Riemannian metric $g$ on the tangent space $T_{\mathbf X}\mathbb{C}^m_n(\mathbf p, \mathbf q)$. For any two tangent vectors $\xi_{\mathbf X}, \eta_{\mathbf X} \in T_{\mathbf X}\mathbb{C}^m_n(\mathbf p, \mathbf q)$, the metric is defined as 
\begin{align}
g(\xi_{\mathbf X}, \eta_{\mathbf X})=\sum_{ij}\frac{(\xi_{\mathbf X})_{ij}(\eta_{\mathbf X})_{ij}}{\mathbf X_{ij}}=\text{Tr}((\xi_{\mathbf X}\oslash \mathbf X)(\eta_{\mathbf X})^T) \label{EqMetric}
\end{align}
where the operator $\oslash$ means the element-wise division of two matrices in the same size.

\begin{remark}
Equivalently we may use the normalized Riemannian metric as follows
\[
g(\xi_{\mathbf X}, \eta_{\mathbf X})=(\mathbf p^T\mathbf{1}_n)\sum_{ij}\frac{(\xi_{\mathbf X})_{ij}(\eta_{\mathbf X})_{ij}}{X_{ij}}.
\]
\end{remark}

As one of building blocks for the optimization algorithms on manifolds, we consider how a matrix of size $n\times m$ can be orthogonally projected onto the tangent space $T_{\mathbf X}\mathbb{C}^m_n(\mathbf p, \mathbf q)$ under its Riemannian metric $g$. 
\begin{theorem}The orthogonal projection from $\mathbb{R}^{n\times m}$ to $T_{\mathbf X}\mathbb{C}^m_n(\mathbf p, \mathbf q)$ takes the following form
\begin{align}
\Pi_{\mathbf X}(\mathbf Y) = \mathbf Y - (\alpha \mathbf{1}^T_m + \mathbf{1}_n \beta^T)\odot \mathbf X,\label{EqProj}  \end{align}
where the symbol $\odot$ denotes the Hadamard product, and $\alpha$ and $\beta$ are given by
\begin{align}
\alpha &= (\mathbf P - \mathbf X\mathbf Q^{-1}\mathbf X)^+(\mathbf Y\mathbf 1_m -\mathbf X\mathbf Q^{-1}\mathbf Y^T\mathbf{1}_n)\in\mathbb{R}^n \label{EqAlpha}\\
\beta &=\mathbf Q^{-1}(\mathbf Y^T\mathbf{1}_n-\mathbf X^T\alpha)\in\mathbb{R}^m \label{EqBeta}
\end{align}
where $\mathbf Z^+$ denotes the pseudo-inverse of $\mathbf Z$, $\mathbf P = \text{diag}(\mathbf p)$ and $\mathbf Q = \text{diag}(\mathbf q)$.
\end{theorem}
\begin{proof} We only present a simple sketch of the proof here. First, it is easy to verify that  for any vectors $\alpha\in\mathbf{R}^n$ and $\beta\in\mathbb{R}^m$, $\mathbf N = (\alpha \mathbf{1}^T_m + \mathbf{1}_n \beta^T)\odot \mathbf X$ is orthogonal to the tangent space $T_{\mathbf X}\mathbb{C}^m_n(\mathbf p, \mathbf q)$. This is because for any $\mathbf S\in T_{\mathbf X}\mathbb{C}^m_n(\mathbf p, \mathbf q)$, we have the following inner product induced by $g$,
\begin{align*}
\langle \mathbf N, \mathbf S\rangle_{\mathbf X} &=\text{Tr}((\mathbf N\oslash\mathbf X)\mathbf S^T)=\text{Tr}((\alpha \mathbf{1}^T_m + \mathbf{1}_n \beta^T)  \mathbf S^T)\\
&=\alpha^T\mathbf S\mathbf{1}_m + \beta^T\mathbf S^T\mathbf 1_n = 0.
\end{align*}

For any $\mathbf Y\in\mathbf{R}^{n\times m}$ and  $\mathbf X\in \mathbb{C}^m_n(\mathbf p, \mathbf q)$, there exist $\alpha$ and $\beta$ such that the following orthogonal decomposition is valid
\[
\mathbf Y = \Pi_{\mathbf X}(\mathbf Y) + (\alpha \mathbf{1}^T_m + \mathbf{1}_n \beta^T)\odot \mathbf X
\]
Hence
\[
\mathbf Y \mathbf{1}_m=  ((\alpha \mathbf{1}^T_m + \mathbf{1}_n \beta^T)\odot \mathbf X)\mathbf{1}_m
\]
By direct element manipulation, we have
\[
\mathbf Y\mathbf 1_m = \mathbf P\alpha + \mathbf X\beta.
\]
Similarly
\[
\mathbf Y^T\mathbf{1}_n = \mathbf X^T\alpha + \mathbf Q\beta.
\]
From the second equation we can express $\beta$ in terms of $\alpha$ as
\[
\beta = \mathbf Q^{-1}(\mathbf Y^T\mathbf 1_n - \mathbf X^T\alpha)
\]
Taking this equation into the first equation gives
\[
\mathbf Y\mathbf{1}_m = (\mathbf P - \mathbf X \mathbf Q^{-1}\mathbf X)\alpha + \mathbf X\mathbf Q^{-1}\mathbf Y^T\mathbf 1_n
\]
This gives both \eqref{EqAlpha} and \eqref{EqBeta}.  The proof is completed.
\end{proof}
\subsection{Riemannian Gradient and Retraction}
The classical gradient descent method can be extended to the case of optimization on manifold with the aid of the so-called Riemannian gradient. As the coupling matrix manifold is embedded in the Enclidean space, the Riemannian gradient can be calculated via projecting the Euclidean gradient onto its tangent space. Given the Riemannian metric which is defined in \eqref{EqMetric}, we can immediately formulate the following lemma, see \cite{SunGaoHongMishraYin2016,DouikHassibi2018},
\begin{lemma}\label{LemmaGrad}
Suppose that $f(\mathbf X)$ is a real-valued smooth function defined on $\mathbb{C}^m_n(\mathbf p, \mathbf q)$ with its Euclidean gradient $\text{Grad}f(\mathbf X)$, then the Riemannian gradient $\text{grad}f(\mathbf X)$ can be calculated as
\begin{align}
\text{grad}f(\mathbf X) = \Pi_{\mathbf X}(\text{Grad}f(\mathbf X)\odot \mathbf X). \label{Eq:7}
\end{align}
\end{lemma}
\begin{proof}As $Df(\mathbf X)[\xi_{\mathbf X}]$, the directional derivative of $f$ along any tangent vector $\xi_{\mathbf X}$, according to the definition of Riemannian gradient, for the metric $g(\cdot, \cdot)$ in \eqref{EqMetric} we have:
\begin{align}
g(\text{grad}f(\mathbf X), \xi_{\mathbf X}) = Df(\mathbf X)[\xi_{\mathbf X}] = \langle \text{Grad}f(\mathbf X), \xi_{\mathbf X}\rangle \label{Eq:8}
\end{align}
where the right equality comes from the definition of Euclidean gradient $\text{Grad}f(\mathbf X)$ with the classic Euclidean metric $\langle\cdot, \cdot\rangle$.  Clearly we have 
\begin{align}
\langle \text{Grad}f(\mathbf X), \xi_{\mathbf X}\rangle = g( \text{Grad}f(\mathbf X)\odot \mathbf X, \xi_{\mathbf X})
\label{Eq:9}
\end{align}
where $g(\text{Grad}f(\mathbf X)\odot \mathbf X, \xi_{\mathbf X})$ can be simply calculated according to the formula in \eqref{EqMetric}, although $\text{Grad}f(\mathbf X)\odot \mathbf X$ is not in the tangent space $T_{\mathbf X}\mathbb{C}^m_n(\mathbf p, \mathbf q)$. Considering its orthogonal decomposition according to the tangent space, we shall have
\begin{align}
\text{Grad}f(\mathbf X)\odot \mathbf X = \Pi_{\mathbf X}(\text{Grad}f(\mathbf X)\odot \mathbf X) + \mathbf Q
\label{Eq:10}
\end{align}
where $\mathbf Q$ is the orthogonal complement satisfying $g(\mathbf Q, \xi_{\mathbf X}) = 0$ for any tangent vector $\xi_{\mathbf X}$. Taking \eqref{Eq:10} into \eqref{Eq:9} and combining it with \eqref{Eq:8} gives
\[
Df(\mathbf X)[\xi_{\mathbf X}] = g(\Pi_{\mathbf X}(\text{Grad}f(\mathbf X)\odot \mathbf X), \xi_{\mathbf X}).
\]
Hence
\[ \text{grad}f(\mathbf X) = \Pi_{\mathbf X}(\text{Grad}f(\mathbf X)\odot \mathbf X).
\]
This completes the proof.
\end{proof}

As an important part of the manifold gradient descent process, the retraction function retracts a tangent vector back to the manifold\cite{AbsilMahonySepulchre2008}. For Euclidean submanifolds, the simplest way to define a retraction is
\[
R_{\mathbf X}(\xi_{\mathbf X}) = \mathbf X + \xi_{\mathbf X}
\]
In our case, to ensure $R_{\mathbf X}(\xi_{\mathbf X})\in \mathbb{C}^m_n(\mathbf p, \mathbf q)$,  $\xi_{\mathbf X}$ should be in the smaller neighbourhood of $\mathbf 0$ particularly when $\mathbf X$ has smaller entries. This will result an inefficient descent optimization process. To provide  a new retraction with high efficiency, following \cite{SunGaoHongMishraYin2016,DouikHassibi2018}, we define $P$  as the projection from the set of element-wise positive matrices $\mathbb{R}^{n\times m}_+$ onto the manifold $\mathbb{C}^m_n(\mathbf p, \mathbf q)$ under the Euclidean metric. Then we have the following lemma. 
\begin{lemma}\label{LemmaProj} For any matrix $\mathbf M\in \mathbb{R}^{n\times m}_+$, there exist two diagonal matrices $\mathbf D_1\in\mathbb{R}^{n\times n}_+$ and $\mathbf D_2\in\mathbb{R}^{m\times m}_+$ such that
\[
P(\mathbf M) = \mathbf D_1 \mathbf M \mathbf D_2 \in \mathbb{C}^m_n(\mathbf p, \mathbf q)
\]
where both $\mathbf D_1$ and $\mathbf D_2$ can be determined by the extended Sinkhorn-Knopp algorithm \cite{PeyreCuturi2019}.
\end{lemma}

The Sinkhorn-Knopp algorithm is specified in Algorithm \ref{Alg1} below, which implements the projection $P$ in Lemma \ref{LemmaProj}. 

\begin{algorithm}
\caption{The Sinkhorn-Knopp Algorithm} \label{Alg1}
\begin{algorithmic}[1] 
\REQUIRE $\mathbf M\in\mathbb{}^{n\times m}_+$, $\mathbf p \in \mathbb{R}^n_+$ and $\mathbf q\in\mathbb{R}^m_+$, a tolerance $\epsilon = 1e-10$ and the number of maximal iteration $T$
\ENSURE $\mathbf D_1$ and $\mathbf D_2$
\STATE Initializing

\hspace{1ex}$\mathbf d_1 = \mathbf q \oslash \mathbf M^T\mathbf 1_m;\;\;\; \mathbf d_2 = \mathbf p \oslash (\mathbf M \mathbf d_1)$;
\WHILE{the iteration is less than $T$}  
\STATE  $\mathbf d_1 = \mathbf q \oslash \mathbf M^T\mathbf d_2;\;\;\; \mathbf d_2 = \mathbf p \oslash (\mathbf M \mathbf d_1);$
\STATE $\mathbf D_1 = \text{diag}(\mathbf d_1)$ and $\mathbf D_2 = \text{diag}(\mathbf d_2);$
\IF{$\|\mathbf D_1\mathbf M\mathbf d_2 - \mathbf p\|<\epsilon$ and $\|\mathbf D_2\mathbf M^T\mathbf d_1 - \mathbf q\|<\epsilon$}
\STATE break while;
\ENDIF 
\ENDWHILE 
\end{algorithmic}
\end{algorithm}

Based on the projection $P$, we define the following retraction mapping for $\mathbb{C}^m_n(\mathbf p, \mathbf q)$
\begin{lemma}\label{Lem6}
Let $P$ be the projection defined in Lemma \ref{LemmaProj}, the mapping $R_{\mathbf X}: T_{\mathbf X}\mathbb{C}^m_n(\mathbf p, \mathbf q)\rightarrow \mathbb{C}^m_n(\mathbf p, \mathbf q)$ given by
\[
R_{\mathbf X}(\xi_{\mathbf X}) = P(\mathbf X\odot \exp(\xi_{\mathbf X}\oslash \mathbf X))
\]
is a valid retraction on  $\mathbb{C}^m_n(\mathbf p, \mathbf q)$.  Here $\exp(\cdot)$ is the element-wise exponential function and $\xi_{\mathbf X}$ is any tangent vector at $\mathbf X$.
\end{lemma}
\begin{proof}
We need to prove that (i) $R_{\mathbf X}(\mathbf 0)=\mathbf X$ and (ii) $\gamma_{\xi_{\mathbf X}}(\tau) = R_{\mathbf X}(\tau\xi_{\mathbf X})$ satisfies $\left.\frac{d\gamma_{\xi_{\mathbf X}}(\tau)}{d\tau}\right|_{\tau=0}=\xi_{\mathbf X}$.

For (i), it is obvious that $R_{\mathbf X}(\mathbf 0)=\mathbf X$ as $P(\mathbf X) = \mathbf X$ for any $\mathbf X\in \mathbb{C}^m_n(\mathbf p, \mathbf q)$.

For (ii), 
\begin{align*}
\left.\frac{d\gamma_{\xi_{\mathbf X}}(\tau)}{d\tau}\right|_{\tau=0} =& \lim_{\tau\rightarrow 0}\frac{\gamma_{\xi_{\mathbf X}}(\tau) - \gamma_{\xi_{\mathbf X}}(0)}{\tau}\\
=&\lim_{\tau\rightarrow 0}\frac{P(\mathbf X\odot \exp(\tau\xi_{\mathbf X}\oslash \mathbf X)) - \mathbf X}{\tau}
\end{align*}
As all $\exp(\cdot)$, $\odot$ and $\oslash$ are element-wise operations, the first order approximation of the exponential function gives \[
P(\mathbf X\odot \exp(\tau\xi_{\mathbf X}\oslash \mathbf X)) = P(\mathbf X + \tau\xi_{\mathbf X}) + o(\tau)
\]
where $\lim_{\tau\rightarrow 0}\frac{o(\tau)}{\tau} = 0$. The next step is to show that $P(\mathbf X + \tau\xi_{\mathbf X})\approx \mathbf X + \tau\xi_{\mathbf X}$ when $\tau$ is very small.  For this purpose, consider a smaller tangent vector $\Delta \mathbf X$ such that $\mathbf X+\Delta\mathbf X \in \mathbb{R}^{n\times m}_+$. There exist two smaller diagonal matrices $\Delta \mathbf D_1\in\mathbb{R}^n_+$ and $\Delta \mathbf D_2\in\mathbb{R}^m_+$ that satisfy
\begin{align*}
P(\mathbf X+\Delta \mathbf X) = (\mathbf I_n + \Delta \mathbf D_1)(\mathbf X+\Delta \mathbf X)(\mathbf I_m + \Delta \mathbf D_2)
\end{align*}
where $\mathbf I$ are identity matrices.  By ignoring higher order small quantity, we have
\[
P(\mathbf X+\Delta \mathbf X) \approx \mathbf X + \Delta\mathbf X + \Delta \mathbf D_1\mathbf X + \mathbf X \Delta \mathbf D_2.
\]
As both $P(\mathbf X+\Delta \mathbf X)$ and $\mathbf X$ are on the coupling matrix manifold and  $\Delta \mathbf X$ is a tangent vector, we have
\begin{align*} 
\mathbf p = & P(\mathbf X+\Delta \mathbf X)\mathbf 1_m \approx (\mathbf X + \Delta\mathbf X + \Delta \mathbf D_1\mathbf X + \mathbf X \Delta \mathbf D_2)\mathbf 1_m\\
\approx & \mathbf p + \mathbf 0 + \Delta \mathbf D_1 \mathbf p + \mathbf X  \Delta \mathbf D_2\mathbf{1}_m  = \mathbf p + \mathbf P \delta \mathbf D_1 + \mathbf X \delta \mathbf D_2
\end{align*}
where $\delta \mathbf D = \text{diag}(\mathbf D)$ and $\mathbf P = \text{diag}(\mathbf P)$. Hence, 
\[
\mathbf P \delta \mathbf D_1 + \mathbf X \delta \mathbf D_2 \approx \mathbf 0.
\]
Similarly,
\[
\mathbf X^T \delta \mathbf D_1 + \mathbf Q \delta \mathbf D_2 \approx \mathbf 0.
\]
That is
\[
\begin{bmatrix} \mathbf P & \mathbf X\\ \mathbf X^T & \mathbf Q\end{bmatrix}\begin{bmatrix}\delta\mathbf D_1 \\ \delta\mathbf D_2\end{bmatrix} \approx \mathbf 0.
\]
Hence $[\delta\mathbf D_1, \delta\mathbf D_2]^T$ is in the null space of the above matrix which contains $[\mathbf 1^T_n, -\mathbf 1^T_m]^T$. In general, there exists a constant $c$ such that $\delta\mathbf D_1 = c\mathbf 1_n$ and $\delta\mathbf D_2 = -c\mathbf 1_m$ and this gives
\[
\Delta\mathbf D_1\mathbf X + \mathbf X \Delta \mathbf D_2 = \mathbf 0.
\]
Combining all results obtained above, we have $P(\mathbf X + \tau\xi_{\mathbf X})\approx \mathbf X + \tau\xi_{\mathbf X}$ as $\tau$ is sufficiently smaller. Hence, this completes the proof.
\end{proof}

\subsection{The Riemannina Hessian}
\begin{theorem}\label{Thm7}
Let $\text{Grad}f(\mathbf X)$ and $\text{Hess}f(\mathbf X)[\xi_{\mathbf X}]$ be the Euclidean gradient and Euclidean Hessian, respectively. The Riemennian Hessian $\text{hess}f(\mathbf X)[\xi_{\mathbf X}]$ can be expressed as
\[
\text{hess}f(\mathbf X)[\xi_{\mathbf X}]=\Pi_{\mathbf X}\left(\dot{\gamma} - \frac12(\gamma\odot \xi_{\mathbf X})\oslash \mathbf X\right)
\]
with
\begin{align*}
\mu =& (\mathbf P - \mathbf X\mathbf Q^{-1}\mathbf X^t)^+\\
\eta =& \text{Grad}f(\mathbf X)\odot \mathbf X\\
\alpha =& \mu (\eta\mathbf 1_m - \mathbf X\mathbf Q^{-1} \eta^T\mathbf{1}_n)\\
\beta =&\mathbf Q^{-1}(\eta^T\mathbf{1}_n - \mathbf X^T\alpha)\\
\gamma =& \eta - (\alpha \mathbf{1}^T_m + \mathbf{1}_n\beta^T)\odot \mathbf X\\
\dot{\mu} =& \mu(\mathbf X \mathbf Q^{-1} \xi^T_{\mathbf X} + \xi_{\mathbf X}\mathbf Q^{-1}\mathbf X^T)\mu\\
\dot{\eta}=& \text{Hess}f(\mathbf X)[\xi_{\mathbf X}]\odot\mathbf X + \text{Grad}f(\mathbf X)\odot \xi_{\mathbf X}\\
\dot{\alpha} =& \dot{\mu}(\eta\mathbf 1_m - \mathbf X\mathbf Q^{-1} \eta^T\mathbf{1}_n)\\
&+\mu(\dot{\eta}\mathbf 1_m -\xi_{\mathbf X}\mathbf Q^{-1} \eta^T\mathbf{1}_n-\mathbf X\mathbf Q^{-1} \dot{\eta}^T\mathbf{1}_n )\\
\dot{\beta}=&\mathbf Q^{-1}(\dot{\eta}^T\mathbf 1_n - \xi_{\mathbf X}^T\alpha - \mathbf X^T\dot{\alpha})\\
\dot{\gamma} = & \dot{\eta} - (\dot{\alpha} \mathbf{1}^T_m + \mathbf{1}_n\dot{\beta}^T)\odot \mathbf X   - (\alpha \mathbf{1}^T_m + \mathbf{1}_n\beta^T)\odot \xi_{\mathbf X}.
\end{align*}
\end{theorem}
\begin{proof}
It is well known \cite{AbsilMahonySepulchre2008} that the Riemannian Hessian can be calculated from the Riemannian connection $\nabla$ and Riemannian gradient via 
\[
\text{hess}f(\mathbf X)[\xi_{\mathbf X}]=\nabla_{\xi_{\mathbf X}}\text{grad}f(\mathbf X).
\]
Furthermore the connection $\nabla_{\xi_{\mathbf X}}\eta_{\mathbf X}$ on the submanifold can be given by the projection of the Levi-Civita connection $\overline{\nabla}_{\xi_{\mathbf X}}\eta_{\mathbf X}$, i.e., $\nabla_{\xi_{\mathbf X}}\eta_{\mathbf X}= \Pi_{\mathbf X}(\overline{\nabla}_{\xi_{\mathbf X}}\eta_{\mathbf X})$. For the Euclidean space $\mathbb{R}^{n\times m}$ endowed with the Fisher information, with the same approach used in \cite{SunGaoHongMishraYin2016}, it can be shown that the Levi-Civita connection is given by
\[
\overline{\nabla}_{\xi_{\mathbf X}}\eta_{\mathbf X}=D(\eta_{\mathbf X})[\xi_{\mathbf X}] - \frac12(\xi_{\mathbf X} \odot \eta_{\mathbf X})\oslash \mathbf X.
\]
Hence,  
\begin{align*}
&\text{hess}f(\mathbf X)[\xi_{\mathbf X}] =  \Pi_{\mathbf X}(\overline{\nabla}_{\xi_{\mathbf X}}\text{grad}f(\mathbf X))   \\
=&\Pi_{\mathbf X}\left(D(\text{grad}f(\mathbf X))[\xi_{\mathbf X}] - \frac12(\xi_{\mathbf X} \odot \text{grad}f(\mathbf X))\oslash \mathbf X\right)
\end{align*}
According to Lemma \ref{LemmaGrad}, the directional derivative can be expressed as
\begin{align*}
&D(\text{grad}f(\mathbf X))[\xi_{\mathbf X}] = D(\Pi_{\mathbf X}(\eta))[\xi_{\mathbf X}]\\
=&D(\eta - (\alpha \mathbf 1^T_m + \mathbf 1_n\beta^T)\odot \mathbf X)[\xi_{\mathbf X}]\\
=&D(\eta)[\xi_{\mathbf X}] - (D(\alpha)[\xi_{\mathbf X}]\mathbf 1^T_m + \mathbf 1_n D(\beta)[\xi_{\mathbf X}]^T)\odot\mathbf X \\
& - (\alpha \mathbf 1^T_m + \mathbf 1_n\beta^T)\odot \xi_{\mathbf X}.
\end{align*}

Taking in the expressions for $\eta, \alpha, \beta$ and directly computing directional derivatives give all formulae in the theorem.
\end{proof}

\section{Riemannian Optimization Applied to OT Problems}
\label{Sec:3}
In this section, we illustrate the Riemannian optimization in solving various OT problems, starting by reviewing the framework of the optimization on Riemannian manifolds. 

\subsection{Optimization on Manifolds}
Early attempts to adapt standard manifold optimization methods were presented by \cite{Gabay1982} in which steepest descent, Newton and qusasi-Newtwon methods 
were introduced. The second-order geometry related optimization algorithm such as the Riemannian trust region algorithm was proposed in \cite{AbsilMahonySepulchre2008}, where the algorithm was applied on some specific manifolds such as the Stiefel and Grassman manifolds.

This paper focuses only on the gradient descent method which is the most widely used optimization method in machine learning.  

Suppose that $\mathbb{M}$ is a $D$-dimensional Riemannian manifold. Let $f: \mathbb{M} \rightarrow \mathbb{R}$ be a real-valued function defined on $\mathbb{M}$. Then, the optimization problem on $\mathbb{M}$ has the form
\[
\min_{\mathbf X\in\mathbb{M}} f(\mathbf X).
\]

For any  $\mathbf X\in\mathbb{M}$ and $\xi_{\mathbf X}\in T_{\mathbf X}\mathbb{M}$, there always exists a geodesic starting at $\mathbf X$ with initial velocity $\xi_{\mathbf X}$, denoted by $\gamma_{\xi_{\mathbf X}}$. With this geodesic the so-called exponential mapping $\exp_{\mathbf X}: T_{\mathbf X}\mathbb{M} \rightarrow \mathbb{M}$ is defined as 
\[
\exp_{\mathbf X}(\xi_{\mathbf X}) = \gamma_{\xi_{\mathbf X}}(1), \;\;\text{for any }\; \xi_{\mathbf X}\in T_{\mathbf X}\mathbb{M}.
\]
Thus the simplest Riemannian gradient descent (RGD) consists of the following two main steps:
\begin{enumerate}
    \item Compute the Riemannian gradient of $f$ at the current position $\mathbf X^{(t)}$, i.e. $\xi_{\mathbf X^{(t)}}=\text{grad}f(\mathbf X^{(t)})$;
    \item Move in the direction $−\xi_{\mathbf X^{(t)}}$ according to $\mathbf X^{(t+1)}=\exp_{\mathbf X^{(t)}}(-\alpha \xi_{\mathbf X^{(t)}})$ with a step-size $\alpha>0$.
\end{enumerate}

Step 1) is straightforward as the Riemannian gradient can be calculated from the Euclidean gradient according to \eqref{Eq:7} in Lemma \ref{LemmaGrad}.  However, it is generally difficult to compute the exponential map effectively as the computational processes require some second-order Riemannian geometrical elements to construct the geodesic, which sometimes is not unique on a manifold point. Therefore, instead of using the exponential map in RGD, an approximated method, namely the retraction map is commonly adopted. 

For coupling matrix manifold $\mathbb{C}^m_n(\mathbf p, \mathbf q)$, a retraction mapping has been calculated in Lemma \ref{Lem6}.  Hence Step 2) in the RGD is defined by
\[
\mathbf X^{(t+1)}=R_{\mathbf X^{(t)}}(-\alpha \xi_{\mathbf X^{(t)}}).
\]

Hence for any given OT-based optimization problem
\[
\min_{\mathbf X\in\mathbb{C}^m_n(\mathbf p, \mathbf q)} f(\mathbf X),
\]
conducting the RGD algorithm comes down to the computation of Euclidean gradient $\text{Grad}f(\mathbf X)$. Similarly, formulating the second-order Riemannian optimization algorithms based on Riemannian Hessian, such as Riemannian Newton method and Riemannian trust region method, 
boil down to calculating 
enable the calcuationn of the Euclidean Hessian. See Theorem \ref{Thm7}.

\subsection{Computational Complexity of Coupling Matrix Manifold Optimization}
In this section we give a simple complexity analysis on optimizing a function defined on the coupling matrix manifold by taking the RGD algorithm as an example. Suppose that we minimize a given objective function $f(\mathbf X)$ defined on $\mathbb{C}^m_n$. For the sake of simplicity, we consider the case of $m=n$. 

In each step of RGD, we first calculate the Euclidean gradient $\text{Grad}f(\mathbf X^{(t)})$ with the number of flops $E_t(n)$. In most cases shown in the next subsection, we have $E_t(n) = O(n^2)$ Before applying the GD step, we shall calculate the Riemannian gradient $\text{grad}f(\mathbf X^{(t)})$ by the projection according to Lemma~\ref{Lem6} which is implemented by the Sinkhorn-Knopp algorithm in Algorithm~\ref{Alg1}. The complexity of Sinkhorn-Knopp algorithm to have an $\epsilon$-approximate solution $O(n\log(n) \epsilon^{-3}) = O(n\log(n))$ \cite{AltschulerWeedRigollet2017}.

If RGD is coducted $T$ iterations, the overall computational complexity will be 
\[
O(n\log(n)T) + TE_t(n) = O(n\log(n)T) + O(Tn^2) = O(Tn^2).
\]

\begin{remark} This complexity is comparable to other optimization algorithms for most OT problems, for example, equivalent to the complexity of  the Order-Preserving OT problem \cite{su2017order}, see Section \ref{Order} below. However as our optimization algorithm has sufficiently exploited the geometry of the manifold, the experimental results are much better than other algorithms, as demonstrated in Section \ref{Sec:4}.
\end{remark}

\begin{remark} Although the Sinkhorn-Knopp algorithm has a complexity of $O(n\log(n))$, it can only be directly applied to solve the entropy regularized OT problem,, see Application Example 2) in Section~\ref{SubSect:C} below.
\end{remark}

\subsection{Application Examples}\label{SubSect:C}
As mentioned before, basic Riemannain optimization algorithms are constructed on the Euclidean gradient and Hessian of the objective function. In the first part of our application example, some classic OT problems are presented to illustrate the calculation process for their Riemannian gradient and Hessian.

\subsubsection{The Classic OT Problem}
The objective function of the classic OT problem \cite{peyre2019computational} is  
\begin{align}
    \min_{\mathbf{X} \in \mathbb{C}^m_n(\mathbf p, \mathbf q)}  f(\mathbf X) = \text{Tr}(\mathbf X^T\mathbf C) \label{EqOld}
\end{align}
where \(\mathbf C = [{C}_{ij}]\in\mathbb{R}^{n\times m}\) is the given cost matrix and $f(\mathbf X)$ gives the overall cost under the transport plan $\mathbf X$. 
The solution $\mathbf X^*$ to this optimization problem is called the transport plan which induces the lowest overall cost $f(\mathbf X^*)$. When the cost is measured by the distance between the source object and the target object, the best transport plan $\mathbf X^*$ assists in defining the so-called Wasserstein distance between the source distribution and the target distribution.  

Given that problem \eqref{EqOld} is indeed a linear programming problem, it is straightforward to solve the problem by the linear programming algorithms. 
In this paper, we solve the OT problem under the Riemannian optimization framework. Thus, for the classic OT, obviously the Euclidean gradient and Hessian can be easily computed as: 
\[
\text{Grad}f(\mathbf X) = \mathbf C
\]
and
\[
\text{Hess}f(\mathbf X)[\xi] = \mathbf 0.
\] 

\subsubsection{The Entropy Regularized OT Problem}   
\par
It is obvious that this classic OT problem can be generalized to the manifold optimization process within our defined coupling matrix manifold \(\mathbb{C}^m_n(\mathbf p, \mathbf q)\) where \(\mathbf p^T\mathbf{1}_n = \mathbf q^T\mathbf{1}_m\) is not necessarily equal to 1, and the number of rows and the number of columns can be unequal.  
To improve the efficiency of the algorithm, we add an entropy regularization term. Hence, the OT problem becomes
\begin{align*}
    \min_{\mathbf{X} \in \mathbb{C}^m_n(\mathbf p, \mathbf q) }f(\mathbf X) = \text{Tr}(\mathbf X^T\mathbf C) - {\lambda}\mathbf{H}(\mathbf{X}), 
\end{align*}
where \(\mathbf{H}(\mathbf{X})\) is the discrete entropy of the coupling matrix and is defined by: 
\begin{align*}
    \mathbf{H}(\mathbf{X}) \triangleq -\sum_{ij} \mathbf{X}_{ij}(\log (\mathbf{X}_{ij})).
\end{align*}
In terms of matrix operation,  $\mathbf H(\mathbf X)$ has the form 
\[
\mathbf{H}(\mathbf{X}) = -\mathbf 1^T_n (\mathbf X\odot \log(\mathbf X))\mathbf{1}_m
\]
where
$\log$ applies to each element of the matrix. The minimization is a strictly convex optimization process, and for \(\lambda >0\) the solution \(\mathbf{X}^*\) is unique and has the form:
 \[
 \mathbf{X}^* =\text{diag}(\boldsymbol{\mu})\mathbf{K}\text{diag}(\boldsymbol{\nu)} 
 \]  where \(\mathbf{K} = e^{\frac{-C}{\lambda}}\) 
 is computed entry-wisely \cite{PeyreCuturi2019}, and $\boldsymbol{\mu}$ and $\boldsymbol{\nu}$ are obtained by the  Sinkhorn-Knopp algorithm. 

Now, for objective function 
\[
f(\mathbf X) = \text{Tr}(\mathbf X^T\mathbf C) - {\lambda} \mathbf{H}(\mathbf X),  
\]
one can easily check that the  Euclidean gradient is
\[
\text{Grad}f(\mathbf X) = \mathbf C + {\lambda} (\mathbb{I}+\log(\mathbf X)),
\]
where $\mathbb{I}$  is a  matrix of all 1s in size $n\times m$, and the Euclidean Hessian is, in terms of mapping differential,  given by
\[
\text{Hess}f(\mathbf X)[\boldsymbol{\xi}] = {\lambda}(\boldsymbol{\xi}\oslash \mathbf X).   
\]

\subsubsection{The Power Regularization for OT Problem}
Dessein \emph{et al.} \cite{DesseinPapadakisRouas2018} further extended the regularization to  
\[
\min_{\mathbf X\in \mathbb{C}^n_n(\mathbf p, \mathbf q)} \text{Tr}(\mathbf X^T\mathbf C) + \lambda \phi(\mathbf X)
\]
where $\phi$ is an appropriate convex function.
As an example, we consider the squared regularization proposed by \cite{EssidSolomon2018}  
\[
\min_{\mathbf X\in \mathbb{C}^n_n(\mathbf p, \mathbf q)} f(\mathbf X) = \text{Tr}(\mathbf X^T\mathbf C) + \lambda \sum_{ij}X^2_{ij}
\]
and we apply a zero truncated operator in the manifold algorithm. It is then straightforward to prove that 
\[
\text{Grad}f(\mathbf X) = \mathbf C + 2\lambda \mathbf X 
\]
and 
\[
\text{Hess}f(\mathbf X)[\boldsymbol{\xi}] = 2\lambda \boldsymbol{\xi}.
\]

The Tsallis Regularized Optimal Transport is used in \cite{MuzellecNockPatriniNielsen2017} to define trot distance which comes with the following regularization problem  
\[
\min_{\mathbf X\in \mathbb{C}^n_n(\mathbf p, \mathbf q)} f(\mathbf X) = \text{Tr}(\mathbf X^T\mathbf C) - \lambda \frac1{1-q}\sum_{ij}(X^q_{ij}  -  X_{ij}).
\]
For the sake of convenience, we denote $\mathbf X^q := [X^q_{ij}]^{n,m}_{i=1,j=1}$ for any given constant $q>0$. Then we have 
\[
\text{Grad}f(\mathbf X) = \mathbf C - \frac{\lambda}{1-q} (q\mathbf X^{q-1} - \mathbb{I}) 
\]
and 
\[
\text{Hess}f(\mathbf X)[\boldsymbol{\xi}] =    q\lambda  \left[\mathbf X^{q-2} \odot\boldsymbol{\xi}\right].
\]

\subsubsection{The Order-Preserving OT Problem}\label{Order} 
The order-preserving OT problem is proposed in \cite{su2017order} and is adopted by \cite{SuWu2019} for learning distance between sequences. This learning process takes the local order of temporal sequences and the learned transport defines a flexible alignment between two sequences. Thus, the optimal transport plan only assigns large loads to the most similar instance pairs of the two sequences.

For sequences $\mathbf U = (\mathbf u_1, ..., \mathbf u_n)$ and $\mathbf V = (\mathbf v_1, ..., \mathbf v_m)$ in the respective given orders,  the distance matrix between them is 
\[
\mathbf C = [d(\mathbf u_i, \mathbf v_j)^2]^{n,m}_{i=1, j=1}.
\]
Define an $n\times m$ matrix (distance between orders)
\[
\mathbf D = \left[\frac1{\left(\frac{i}n - \frac{j}m\right)^2+1}\right]
\]
and the (exponential) similarity matrix 
\[
\mathbf P = \frac1{\sigma\sqrt{2\pi}}\left[\exp\left\{-\frac{l(i,j)^2}{2\sigma^2}\right\}\right]
\]
where $\sigma>0$ is the scaling factor and 
\[
l(i,j) = \left|\frac{\frac{i}n - \frac{j}m}{\sqrt{\frac1{n^2} + \frac1{m^2}}}\right|.
\]
The (squared) distance between sequences $\mathbf U$ and $\mathbf V$ is given by  
\begin{align}
d^2(\mathbf U, \mathbf V) = \text{Tr}(\mathbf C^T\mathbf X^*) \label{Eq:Dist}
\end{align}
where the optimal transport plan $\mathbf X^*$ is the solution to the following order-preserving regularized OT  
problem  
\[
\mathbf X^* = \argmin_{\mathbf X\in \mathbb{C}^m_n(\mathbf p, \mathbf q)} f(\mathbf X) = \text{Tr}(\mathbf X^T(\mathbf C - \lambda_1 \mathbf D)) {+} \lambda_2  \text{KL}(\mathbf X||\mathbf P)
\]
where the KL-divergence is defined as
\[
\text{KL}(\mathbf X||\mathbf P) = \sum_{ij}X_{ij}(\log(X_{ij}) - \log(P_{ij}))
\]
and specially $\mathbf p = \frac1n\mathbf 1_n$ and $\mathbf q = \frac1m\mathbf 1_m$ are uniform distributions. 
Hence
\[
\text{Grad}f(\mathbf X) = (\mathbf C - \lambda_1 \mathbf D) {+} \lambda_2 (\mathbb{I} + \log (\mathbf X) - \log (\mathbf P)) 
\]
and 
\[
\text{Hess}f(\mathbf X)[\boldsymbol{\xi}] = \lambda_2 (\boldsymbol{\xi}\oslash \mathbf X).
\]

\subsubsection{The OT Domain Adaption Problem}
OT has also been widely used for solving the domain adaption problems.
In this subsection, the authors of \cite{courty2016optimal} formalized two class-based regularized OT problems, namely the group-induced OT (OT-GL) and the Laplacian regularized OT (OT-Laplace). As the OT-Laplace is found to be the best performer for domain adaption, we only apply our coupling matrix manifold optimization to it  and thus we summarize its objective function here.

As pointed out in \cite{courty2016optimal}, this regularization aims at preserving the data graph structure during transport. Consider $\mathbf P_s = [\mathbf p^s_{1}, \mathbf p^s_{2}, ..., \mathbf p^s_{n}]$ to be the $n$ source data points and $\mathbf P_t = [\mathbf p^t_{1}, \mathbf p^t_{2}, ..., \mathbf p^t_{m}]$ the $m$ target data points, both are defined in $\mathbb{R}^d$. Obviously,   $\mathbf P_s\in\mathbb{R}^{d\times n}$ and $\mathbf P_t\in\mathbb{R}^{d\times m}$.  The purpose of domain adaption is to transport the source $\mathbf P_s$ towards the target $\mathbf P_t$ so that the transported source $\widehat{\mathbf P}_s =[\widehat{\mathbf p}^s_{1}, \widehat{\mathbf p}^s_{2}, ..., \widehat{\mathbf p}^s_{n}]$ and the target $\mathbf P_t$ can be jointly used for other learning tasks. 

Now suppose that for the source data we have extra label information $\mathbf Y_s = [y^s_1, y^2_2, ..., y^s_n]$. With this label information we sparsify similarities $\mathbf S_s=[S_s(i,j)]^n_{i,j=1} \in\mathbb{R}^{n\times n}_+$ among the source data such that $S_s(i,j) = 0$ if $y^s_i \not= y^s_j$ for $i,j=1,2, ..., n$. That is, we define a $0$ similarity between two source data points if they do not belong to the same class or do not have the same labels.  Then the following regularization is proposed
\begin{align*}
\Omega^s_c(\mathbf X) = \frac1{n^2}\sum^n_{i,j=1}S_s(i,j)\|\widehat{\mathbf p}^s_{i}-\widehat{\mathbf p}^s_{j}\|^2_2. 
\end{align*}

With a given transport plan $\mathbf X$, we can use the barycentric mapping in the target as the transported point for each source point \cite{courty2016optimal}. When we use the uniform marginals for both source and target and the $\ell_2$ cost, the transported source is expressed as 
\begin{align}
\widehat{\mathbf P}_s = n \mathbf X \mathbf{P}_t. \label{TransPorted}
\end{align}
It is easy to verify that
\begin{align}
\Omega^s_c(\mathbf X) = \text{Tr}(\mathbf{P}^T_t \mathbf{X}^T \mathbf{L}_s \mathbf{X}\mathbf{P}_t),
\label{LapReg}
\end{align}
where $\mathbf{L}_s = \text{diag}(\mathbf{S}_s\mathbf{1}_n) - \mathbf{S}_s$ is  the  Laplacian  of  the graph $\mathbf{S}_s$ and the regularizer $\Omega_c(\mathbf X)$ is therefore quadratic with respect to $\mathbf X$.  Similarly when the Laplacian $\mathbf L_t$ in the target domain is available, the following symmetric Laplacian regularization is proposed
\begin{align*}
\Omega_c(\mathbf X) & =  (1-\alpha)\text{Tr}(\mathbf{P}^T_t \mathbf{X}^T \mathbf{L}_s \mathbf{X}\mathbf{P}_t) + \alpha \text{Tr}(\mathbf{P}^T_s \mathbf{X} \mathbf{L}_t \mathbf{X}^T\mathbf{P}_s)\notag \\
&= (1-\alpha)\Omega^s_c(\mathbf X) + \alpha \Omega^t_c(\mathbf X).
\end{align*}
When $\alpha = 0$, this goes back to the regularizer $\Omega^s_c(\mathbf X)$ in \eqref{LapReg}. 

Finally the OT domain adaption is defined by the following Laplacian regularized OT problem
\begin{align}
\min_{\mathbf X\in\mathbb{C}^m_n(\mathbf 1_n, \mathbf 1_m)}f(\mathbf X) = \text{Tr}(\mathbf X^T\mathbf C) - {\lambda} \mathbf{H}(\mathbf X) + \frac12\eta \Omega_c(\mathbf X) \label{Eq:Symm}
\end{align}
Hence the Euclidean gradient and uclidean Hessian are given by 
\begin{align*}
\text{Grad}f(\mathbf X) = & \mathbf C + {\lambda} (\mathbb{I}+ \log(\mathbf X)) \\
& + \eta ((1-\alpha)\mathbf{L}_s \mathbf X\mathbf{P}_t\mathbf{P}^T_t + \alpha \mathbf P_s\mathbf P^T_s \mathbf X \mathbf L_t).
\end{align*}
and 
\[
\text{Hess}f(\mathbf X)[\boldsymbol{\xi}] = {\lambda}(\boldsymbol{\xi}\oslash \mathbf X) + \eta ((1-\alpha) \mathbf{L}_s \boldsymbol{\xi}\mathbf{P}_t\mathbf{P}^T_t + \alpha \mathbf P_s\mathbf P^T_s \boldsymbol{\xi} \mathbf L_t), 
\]  
respectively.

\section{Experimental Results and Comparisons}\label{Sec:4}
In this section, we investigate the performance of our proposed methods. The implementation of the coupling matrix manifold follows the framework of ManOpt Matlab toolbox in \url{http://www.manopt.org} from which we call the conjugate gradient descent algorithm as our Riemannian optimization solver in experiments.  All experiments are carried out on a laptop computer running on a 64-bit operating system with Intel Core i5-8350U 1.90GHz CPU and 16G RAM with MATLAB 2019a version.

\subsection{Synthetic Data for the Classic OT Problem}
First of all, we conduct a numerical experiment on a classic OT problem with synthetic data and the performance of the proposed optimization algorithms are demonstrated. 

Consider the following source load $\mathbf p$ and target load $\mathbf q$, and their per unit  cost matrix $\mathbf C$:
\begin{align*}
\mathbf p = \begin{bmatrix}3 \\ 3 \\ 3 \\ 4 \\ 2 \\ 2 \\ 2 \\ 1\end{bmatrix},
\;\;\mathbf q = \begin{bmatrix}4 \\ 2 \\ 6 \\ 4 \\ 4 \end{bmatrix},\;\;\mathbf C = \begin{bmatrix}
0 &	0&	1.2&	2&	2\\
2&	4&	4&	4&	0\\
1&	0&	0&	0&	3\\
0&	1&	2&	1&	3\\
1&	1&	0&	1&	2\\
2&	1&	2&	0.8& 3\\
4&	0&	0&	1&	1\\
0&	1&	0&	1&	3
\end{bmatrix}.
\end{align*}

\begin{figure}[tbh]
\centering
\subfloat[LinProg]{\includegraphics[width =0.32\textwidth]{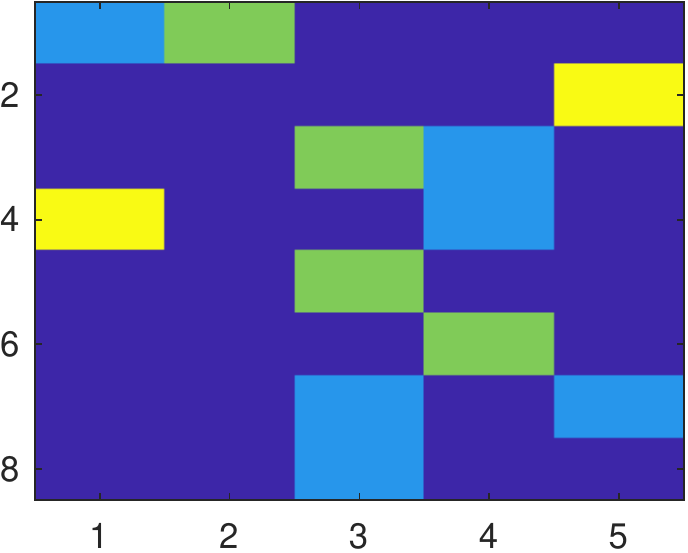}}\;\;\;
\subfloat[CMM]{\includegraphics[width =0.32\textwidth]{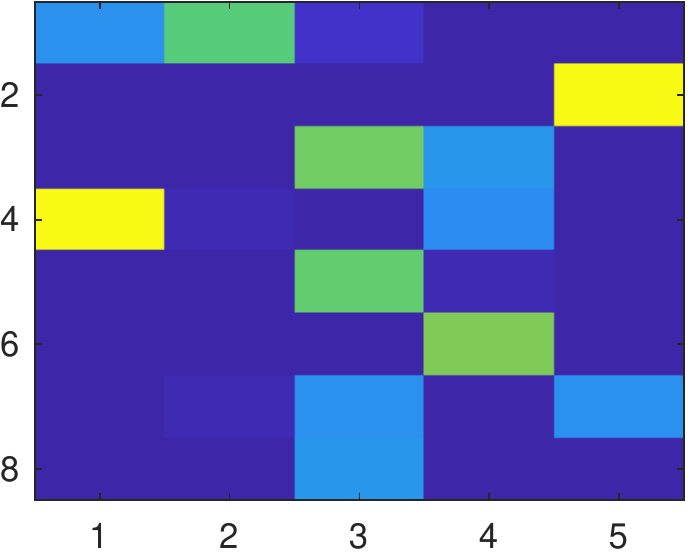}}
\caption{Two transport plan matrices via: (a) Linear Programming and (b) Coupling Matrix Manifold Optimization.}\label{Figure1}
\end{figure}

For this setting, we solve the classic OT problem using the coupling matrix manifold optimization (CMM) and the standard linear programming (LinProg) algorithm, respectively. We visualize the learned transport plan matrices from both algorithms in Fig.~\ref{Figure1}. 
The results reveal that the linear programming algorithm is constrained by a non-negative condition for the entries of transport plan and hence the output transportation plan demonstrates the sparse pattern. While our coupling matrix manifold imposes the positivity constraints, it generates a less sparse solution plan, which give a preferred pattern in many practical problems. The proposed manifold optimization perform well in this illustrative example.

Next we consider an entropy regularized OT problem which can be easily solved by the Sinkhorn algorithm. We test both the Sinkhorn algorithm and the new coupling matrix manifold optimization on the same synthetic problem over 100 regularizer $\lambda$ values on a log scale ranging $[-2, 2]$, i.e., $\lambda = 0.001$ to $100.0$. Mean squared error is used as a criterion to measure the closeness between transport plan matrices in both algorithms.  
\begin{figure}[tbh]
\centering
\includegraphics[width =0.38\textwidth]{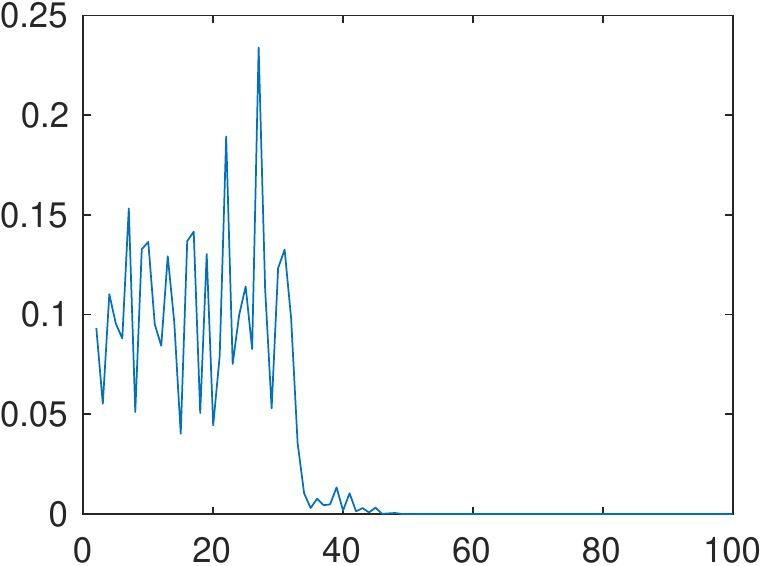}
\caption{The error between two transport plan matrices given by two algorithms verse the regularizer $\lambda$.}\label{Figure2}
\end{figure}

From Fig.~\ref{Figure2}, we observe that when the Sinkhorn algorithm breaks down for $\lambda<0.001$  due to computational instability. On the contrary, the manifold-assisted algorithm generates reasonable results for a  wider range of regularizer values. We also observe that both algorithms give almost exactly same transport plan matrices when $\lambda>0.1668$. However, in terms of computational time, the Sinkhorm algorithm is generally more efficient than the manifold assisted method in the entropy regularize OT problem,

\subsection{Experiments on the Order-Preserving OT}
In this experiment, we demonstrate the performance in calculating the order-preserving Wasserstein distance \cite{su2017order} using a real dataset. The ``Spoken Arabic Digits (SAD)'' dataset, available from the UCI Machine Learning Repository (\url{https://archive.ics.uci.edu/ml/datasets/Spoken+Arabic+Digit}), contains 8,800 vectorial sequences from ten spoken Arabic digits. The sequences consist of time series of the mel-frequency cepstrumcoefficients (MFCCs) features extracted from the speech signals. This is a classification learning task on ten classes. The full set of training data has 660 sequence samples per digit spoken repeatedly for 10 times by 44 male and 44 female Arabic native speakers. For each digit, another 220 samples are retained as testing sets. 

The experimental setting is similar to that in \cite{su2017order}. Based on the order-preserving Wasserstein distance (OPW) between any two sequence, we directly test the nearest neighbour (NN) classifier.  To define the distance in \eqref{Eq:Dist}, we use three hyperparameters: the width parameter $\sigma$ of the radius basis function (RBF), two regularizers $\lambda_1$ and $\lambda_2$. For the comparative purpose, these hyperparameters are chosen to be $\sigma =1$, $\lambda_1 =50$ and $\lambda_2 = 0.1$, as in \cite{su2017order}. Our purpose here is to illustrate that the performance of the NN classifier based on the coupling matrix manifold optimization algorithm (named as CM-OPW) is comparable to the NN classification results from Sinkhorn algorithm (named as S-OPW). We randomly choose 10\% training data and 10\% testing data for each run in the experiments. The classification mean accuracy and their standard error are reported in TABLE~\ref{Table2} based on five runs. 
\begin{table}
    \centering
    \begin{tabular}{|c|cccccc|}\hline
    Algorithms & 1NN & 3NN & 5NN & 7NN & 13NN & 19NN\\ \hline
   S-OWP \cite{su2017order} & 0.8236 & 0.8454 &  0.8454 &  
0.8418 & 0.8473 &  0.8290  \\
(std)  & 0.0357 & 0.0215& 0.0215& 0.0220& 0.0272& 0.0240\\ \hline
CM-OWP &  0.8091 &0.8309 & 0.8255 &0.8218 & 0.8109 &0.8091\\
(std) &0.0275 & 0.0212 &0.0194 &0.0196& 0.0317 &0.0315 \\
    \hline
    \end{tabular}
    \caption{The classification accuracy of the kNN classifiers based on two algorithms for the order-preserving Wasserstein distance. }
    \label{Table2}
\end{table}

In this  experiment, we also observe that the distance calculation fails for some pairs of training and testing sequences due to numerical instability of the Sinkhorn algorithm. Our conclusion is that the performance of the manifold-based algorithm is comparable in terms of similar classification accuracy.  When $k= 1$, the test sequence is also viewed as a query to retrieve the training sequences, and the mean average precision (MAP) is  $\text{MAP} = 0.1954$ for the S-OPW and $\text{MAP} = 0.3654$ for CM-OPW. Theoretically the Sinkhorn algorithm is super-fast, outperforming all other existing algorithms; however, it is not applicable to those OT problems with non-entropy regularizations. We demonstrate these problems in the next subsection.

\subsection{Laplacian Regularized OT Problems: Synthetic Domain Adaption}
Courty \emph{et al.}  \cite{courty2016optimal} analyzed two moon datasets and found that the OM domain adaption method significantly outperformed the subspace alignment method significantly. 
\begin{figure}[tbh]
\centering
\subfloat[rotation = 10\degree]{\includegraphics[width =0.22\textwidth, height =4cm]{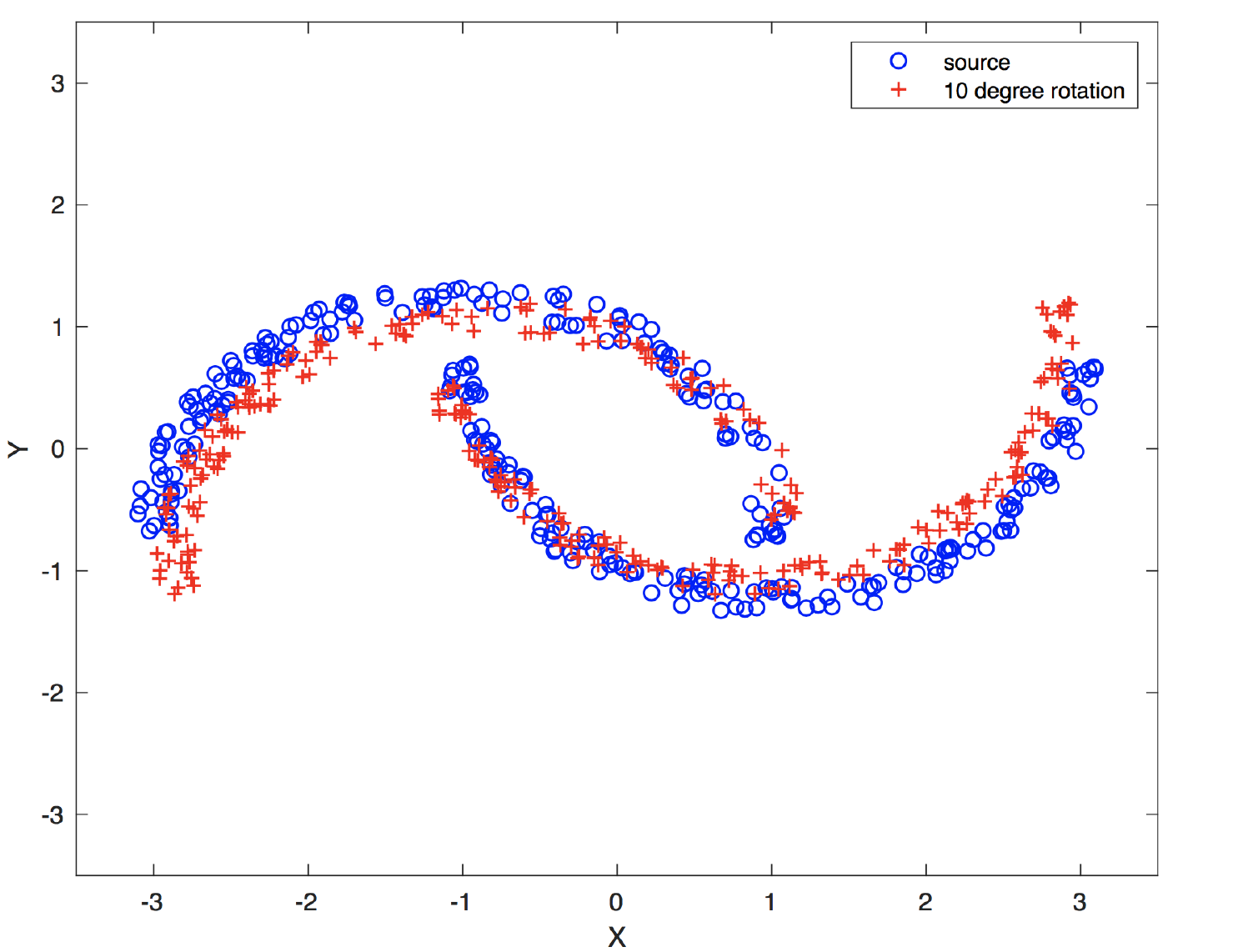}}\;\;\;
\subfloat[rotation = 30\degree]{\includegraphics[width =0.22\textwidth,height =4cm]{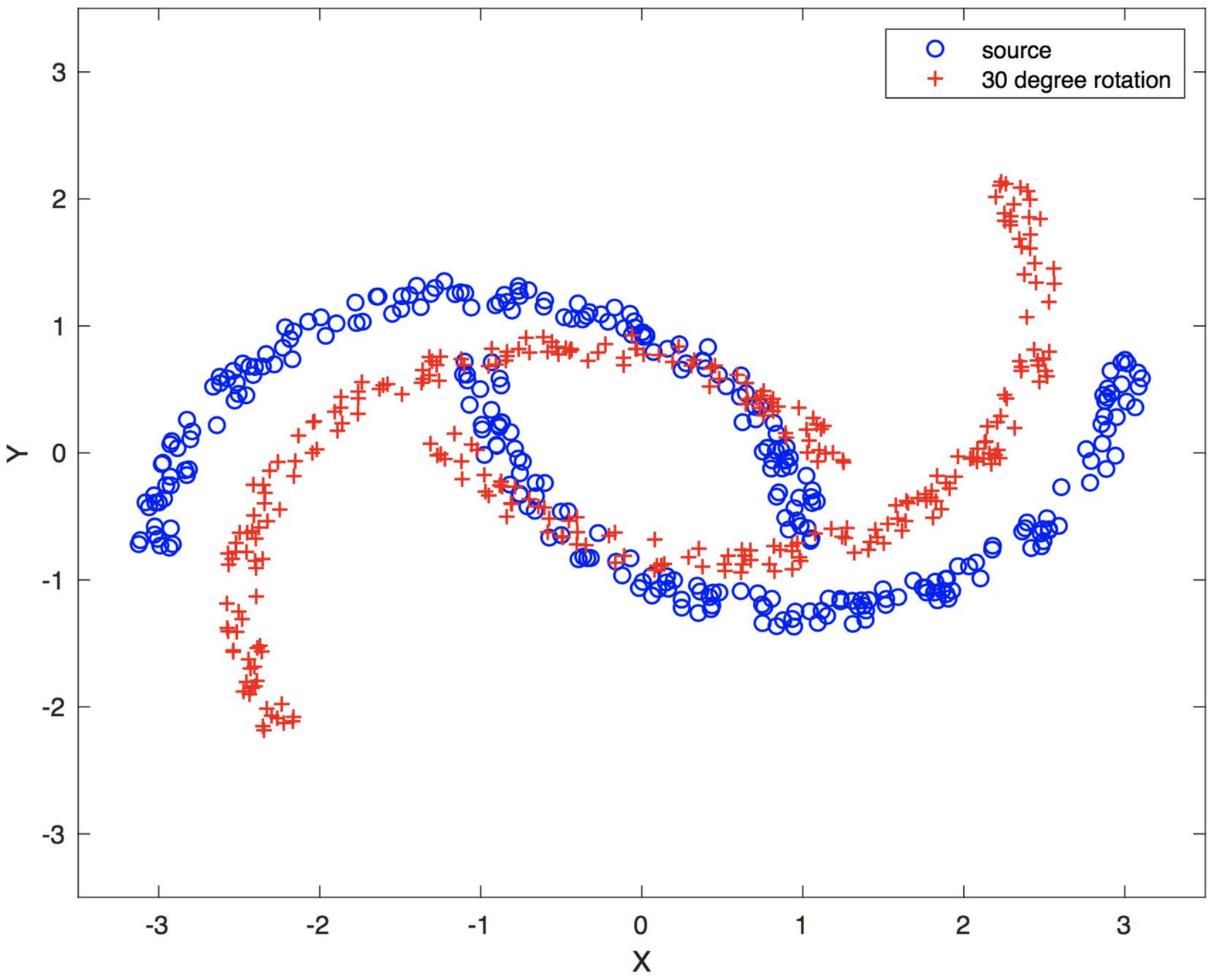}}\;\;\;
\subfloat[rotation = 50\degree]{\includegraphics[width =0.22\textwidth,height =4cm]{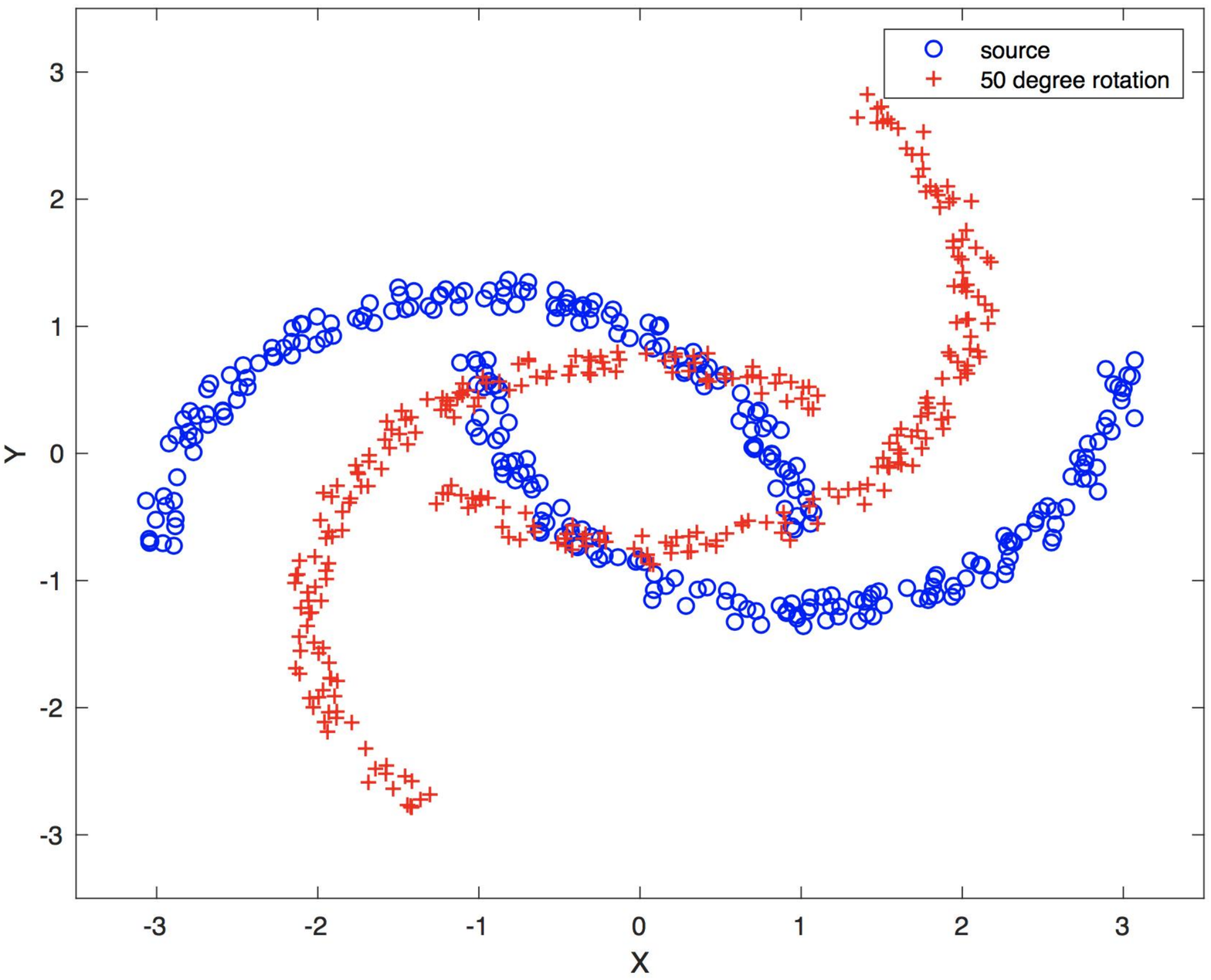}}\;\;\;
\subfloat[rotation = 90\degree]{\includegraphics[width =0.22\textwidth,height =4cm]{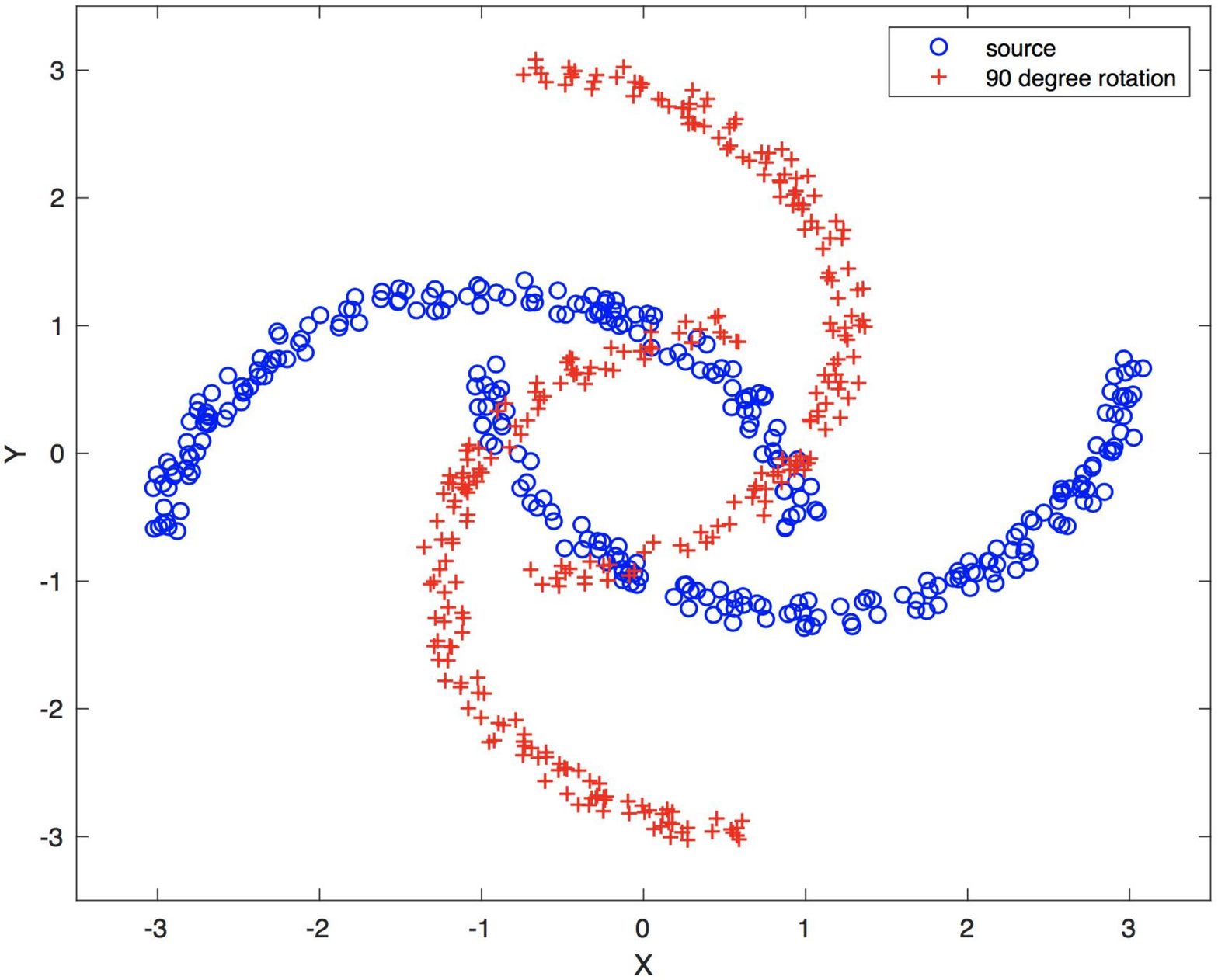}}
\caption{Two moons' example for increasing rotation angles }\label{Figure3}
\end{figure}
We use the same experimental data and protocol as in  \cite{courty2016optimal} to perform a direct and fair comparison between results\footnote{We sincerely thanks to the authors of \cite{courty2016optimal} for providing us the complete simulated two moon datasets.}. Each of the two domains represents the source and the target respectively presenting two moon shapes associated with two specific classes. See Fig.~\ref{Figure3}.

The  source  domain contains 150 data points sampled from the two moons. Similarly, the  target  domain has the same number of data points, sampled  from two moons shapes which rotated at a given angle from the base moons used in the source domain. A classifier between the data points from two domains will be trained once transportation process is finished. 

To test the generalization capability of the classifier based on the manifold optimization method, we sample a set of 1000 data points according to the distribution of the target domain and we repeat the experiment for 10 times, each of which is conducted on 9 different target domains corresponding to $10\degree$, $20\degree$, $30\degree$, $40\degree$, $50\degree$, $60\degree$, $70\degree$, $80\degree$ and $90\degree$ rotations, respectively.  We report the mean classification error and variance as comparison criteria.  

We  train  the SVM classifiers with a Gaussian kernel, whose parameters were automatically set by 5-fold cross-validation. The final results are shown in TABLE~\ref{Table1}. For comparative purpose, we also present the results based on the DA-SVM approach \cite{BruzzoneMarconcini2010} and the PBDA  \cite{GermainHabrardLavioletteMorvant2013} from \cite{courty2016optimal}. 
\begin{table*}
    \centering
    \begin{tabular}{|c|ccccccc|}\hline
    Rotate Angle     &  $10\degree$ & $20\degree$ & $30\degree$ & $40\degree$ & $50\degree$ &   $70\degree$ & $90\degree$\\ \hline
   SVM (no adapt.) & \textbf{0}    & 0.104   & 0.24   & 0.312    & 0.4  &  0.764  & 0.828\\
   DASVM  &\textbf{0}    &  \textbf{0}  &   0.259 &  0.284  & 0.334  & 0.747  & 0.82\\
   PBDA & \textbf{0}   & 0.094 &  0.103 &  0.225 &  0.412 &  0.626  & 0.687 \\
   OT-Laplace  & \textbf{0}& \textbf{0}& 0.004 & 0.062& 0.201 & 0.402 & 0.524\\
   \hline
   \textbf{CM-OT-Lap (ours)} & 0.0027  &  0.0043   & \textbf{0.0014}  &  \textbf{0.0142}  &  \textbf{0.0301}  &   \textbf{0.0446}  &  \textbf{0.0797}\\ 
   \textbf{(variance)} & 0.0000  &   0.0002  &   0.0000  &  0.0007  &   0.0013   &     0.0015    & 0.0057 \\ \hline
    \end{tabular}
    \caption{Mean error rate over 10 realizations for the two moons simulated example. DASVM \cite{BruzzoneMarconcini2010}; PBDA \cite{GermainHabrardLavioletteMorvant2013}; OT-Laplace \cite{courty2016optimal}  }
    \label{Table1}
\end{table*}

From TABLE~\ref{Table1}, we observe that the coupling matrix manifold assisted optimization algorithm significantly improves the efficiency of the GCG (the generalized conditional gradient) algorithm which ignores the manifold constraints while a weaker Lagrangian condition was imposed in the objective function. This results in a sub-optimal solution to the transport plan, producing poorer transported source data points.

\subsection{Laplacian Regularized OT Problems: Image Domain Adaption}
We now apply our manifold-based algorithm to solve the Laplician regularized OT problem for the challenging real-world adaptation tasks. In this experiment, we test the domain adaption for both handwritten digits images and face images for recognition. We follow the same setting used in \cite{courty2016optimal} for a fair comparison. 

\subsubsection{\textbf{Digit recognition}}
We use the two-digit famous handwritten digit datasets USPS and MNIST as the source and target domain and verse, respectively, in our experiment\footnote{Both datasets can be found at \url{http://www.cad.zju.edu.cn/home/dengcai/Data/MLData.html}.}.  The datasets share 10 classes of features (single digits from 0-9). We randomly sampled 1800 images from USPS and 2000 from MNIST. In order to unify the dimensions of two domains, the MNIST images are re-sized into \(16\times16\) resolution same as USPS. The grey level of all images are then normalized to produce the final feature space for all domains. For this case, we have two settings U-M (USPS as source and MNIST as target) and M-U (MNIST as source and USPS as target).

\subsubsection{\textbf{Face Recognition}}

In the face recognition experiment, we use PIE (``Pose, Illumination, Expression'') dataset which contain \(32 \times 32\) images of 68 individuals with different poses: pose, illuminations and expression conditions\footnote{\url{http://www.cs.cmu.edu/afs/cs/project/PIE/MultiPie/Multi-Pie/Home.html}}. In order to make a fair and reasonable comparison with \cite{courty2016optimal}, we select PIE05(C05, denoted as P1, left pose), PIE07(C07, denote as P2, upward pose), PIE09(C09, denoted as P3, downward pose) and PIE29(C29, denoted as P4, right pose). This four domains induce 12 adaptation problems with increasing difficulty (the hardest adaptation is from left to the right). Note that large variability between each domain is due to the illumination and expression. 

\subsubsection{\textbf{Experiment Settings and Result Analysis}}

We generate the experimental results by applying the manifold-based algorithm on two types of Laplacian regularized problems, namely: Problem \eqref{Eq:Symm} with $\alpha = 0$ (CMM-OT-Lap) and with $\alpha = 0.5$ (CMM-OT-symmLap). We follow the same experimental settings in \cite{courty2016optimal}. For all methods, the regularization parameter \(\lambda\) was initially set to 0.01， similarly, another parameter,  \(\eta\) that controls the performance of Laplacian terms was set to 0.1.  

In both Face and digital recognition experiments, 1NN is trained with the adapted source data and target data, and then we report the overall accuracy (OA) score (in \%) calculated on testing samples from the target domain.   We compare OAs between our CMM-OT solutions to the baseline methods and the results generated by the methods provided in \cite{courty2016optimal} in TABLE~\ref{Table3}. Note that, we applied both coupling matrix OT Laplacian and coupling matrix OT symmetric Laplacian algorithm for all experiments, and due to the high similarity of the results generated from these two methods, we only list the OA generated from the non-symmetric CMM-OT-Lap algorithm in table.

As a result, the OA based on the solution generated from CMM based OT Laplician algorithm over-performs all other methods in both digital and face recognition experiments, with mean OA = \(65.52\%\) and \(72.59\%\), respectively. Averagely, our method is able to increase 4\% and 16\% of the OA from the previous results. However, in terms of the adaptation problem with the highest difficulty : P1 to P4, we got similar result compared with previous results, with the OA = \(47.54\%\) from \cite{courty2016optimal} and \(48.98\%\) from our method respectively. 
\begin{table}[htbp]
  \centering
  \begin{tabular}{|c|ccc|c|} \hline
    Domains  & \textbf{1NN}   & \textbf{OT-IT} & \textbf{OT-Lap} & \textbf{CMM-OT-Lap} \\\hline
   U-M  & 39.00 &53.66 &57.43 & \textbf{60.67} \\ 
      M-U  &58.33  &64.73 &64.72 & \textbf{70.37} \\  
    mean &48.66 &59.20  &61.07 & \textbf{65.52}  \\ \hline
    P1-P2 & 23.79 & 53.73 & 58.92 & \textbf{58.08} \\
    P1-P3 & 23.50 & 57.43 & 57.62 & \textbf{62.65} \\
    P1-P4 & 15.69 & 47.21 & 47.54 & \textbf{48.98} \\
    P2-P1 & 24.27 & 60.21 & 62.74 & \textbf{93.10} \\
    P2-P3 & 44.45 & 63.24 & 64.29 & \textbf{69.18} \\
    P2-P4 & 25.86 & 51.48 & 53.52 & \textbf{65.10} \\
    P3-P1 & 20.95 & 57.50 & 57.87 & \textbf{91.70} \\
    P3-P2 & 40.17 & 63.61 & 65.75 & \textbf{75.66} \\
    P3-P4 & 26.16 & 52.33 & 54.02 & \textbf{87.60} \\
    P4-P1 & 18.14 & 45.15 & 45.67 & \textbf{90.30} \\
    P4-P2 & 24.37 & 50.71 & 52.50 & \textbf{66.46} \\
    P4-P3 & 27.30 & 52.10 & 52.71 & \textbf{62.29} \\
    mean  & 26.22 & 54.56 & 56.10 & \textbf{72.59} \\ \hline
    \end{tabular} 
    \caption{Overall recognition accuracies in \% in both digital and face recognition} 
  \label{Table3}%
\end{table}%

\section{Conclusions}\label{Sec:5}

This paper explores the so-called coupling matrix manifolds on which the majority of the OT objective functions are defined. We formally defined the manifold, explored its tangent spaces, defined a Riemennian metric based on information measure, proposed all the formulas for the Riemannian gradient, Riemannina Hessian and an appropriate retraction as the major ingradients for implementation Riemannian optimization on the manifold. We apply manifold-based optimization algorithms (Riemannian gradient descent and second-order Riemannian trust region)
into several types of OT problems, including the classic OT problem, the entropy regularized OT problem, the power regularized OT problem, the state-of-the-art order-preserving Wasserstein distance problems and the OT problem in regularized domain adaption applications. The results from three sets of numerical experiments demonstrate that the newly proposed Riemannian optimization algorithms perform as well as the classic algorithms such as Sinkhorn algorithm. We also find that the new algorithm overperforms the generalized conditional gradient when solving non-entropy regularized OT problem where the classic Sinkhorn algorithm is not applicable. 

\section*{Acknowledgement}
This project is partially supported by the University of Sydney Business School ARC Bridging grant.

\bibliographystyle{spbasic}
\bibliography{refer} 
\end{document}